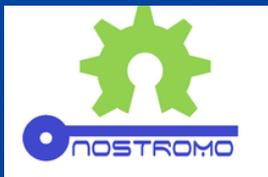

NEXT-GENERATION OPEN-SOURCE TOOLS FOR ATM PERFORMANCE MODELLING AND OPTIMISATION

# White Paper

# NOSTROMO: Lessons learned, conclusions and way forward

This project has received funding from the SESAR Joint Undertaking (SJU) under grant agreemet Nº 892517. The SJU receives support from the European Union's Horizon 2020 research and innovation programme and the SESAR JU members other than the Union. © 2020 NOSTROMO Consortium. All rights reserved.

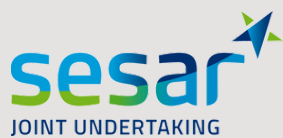
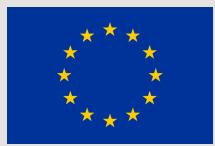

# NOSTROMO - LESSONS LEARNED AND CONCLUSIONS

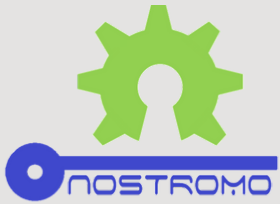

**NEXT-GENERATION OPEN-SOURCE TOOLS FOR ATM PERFORMANCE MODELLING AND OPTIMISATION**

**Authors:**

Mayte Cano (CRIDA)
Andrés Perillo (CRIDA)
Juan Antonio López (CRIDA)
Faustino Tello (CRIDA)
Javier Poveda (CRIDA)
Francisco Câmara (DTU)
Francisco Antunes (DTU)
Christoffer Riis (DTU)
Ian Crook (ISA)
Abderrazak Tibichte (ISA)
Sandrine Molton (ISA)
David Mocholí (Nommon)
Ricardo Herranz (Nommon)
Gérald Gurtner (UoW)
Tatjana Bolić (UoW)
Andrew Cook (UoW)
Jovana Kuljanin (UPC)
Xavier Prats (UPC)

**Cooperation & Funding:**

This project has received funding from the SESAR3 Joint Undertaking (SJU) under grant agreement No 892517. The SJU receives support from the European Union's Horizon 2020 research and innovation programme.

The main objective of the NOSTROMO project was to develop, demonstrate and evaluate an innovative modelling approach for the rigorous and comprehensive assessment of the performance impact of future ATM concepts and solutions at ECAC network level. The activities proposed by NOSTROMO range from basic research to applied research, from TRL1 to TRL2.

Read more: https://nostromo-h2020.eu/

**Disclamer:**

All images used in this brochure are either from NOSTROMO or free stock photos.

**Partners:**

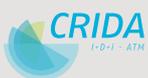 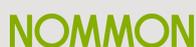 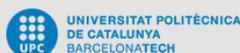 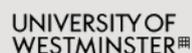 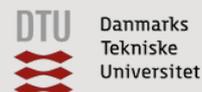 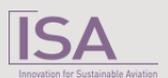

# NOSTROMO: Lessons learned, conclusions and way forward

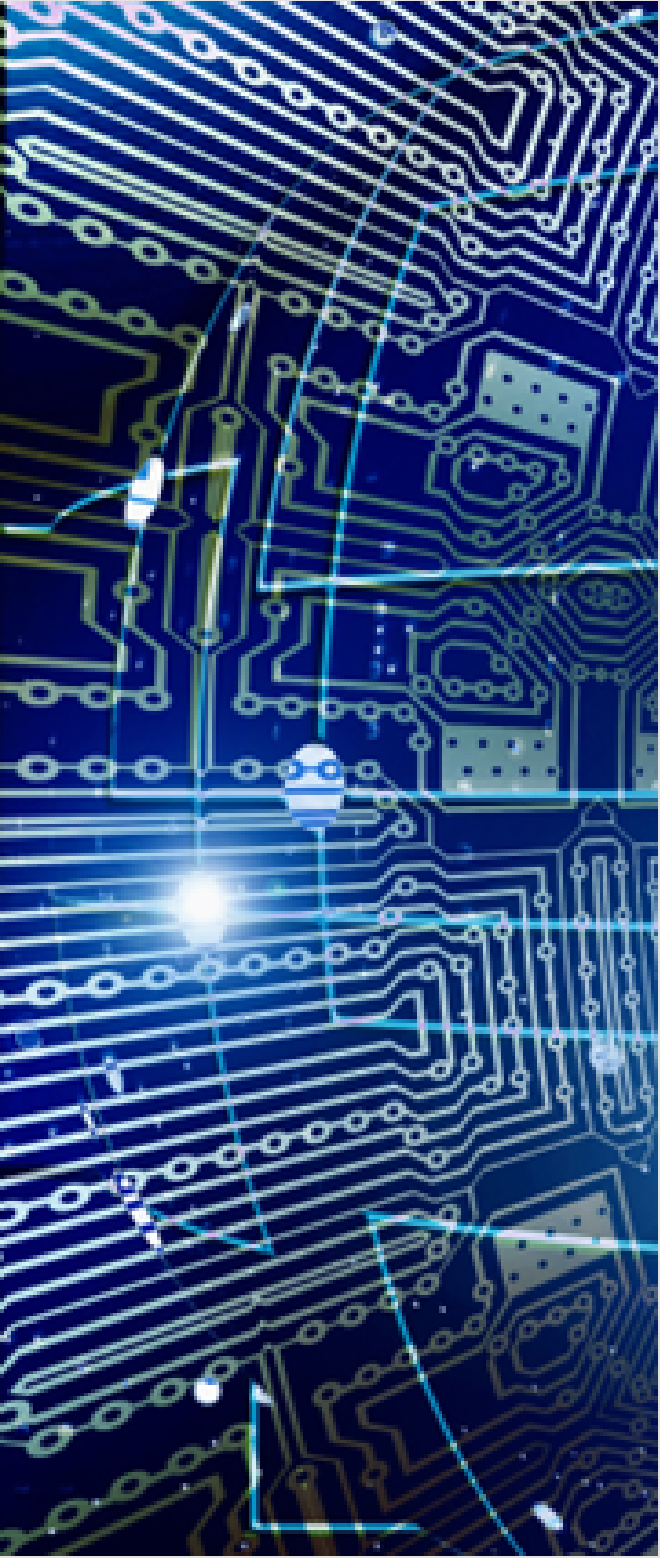

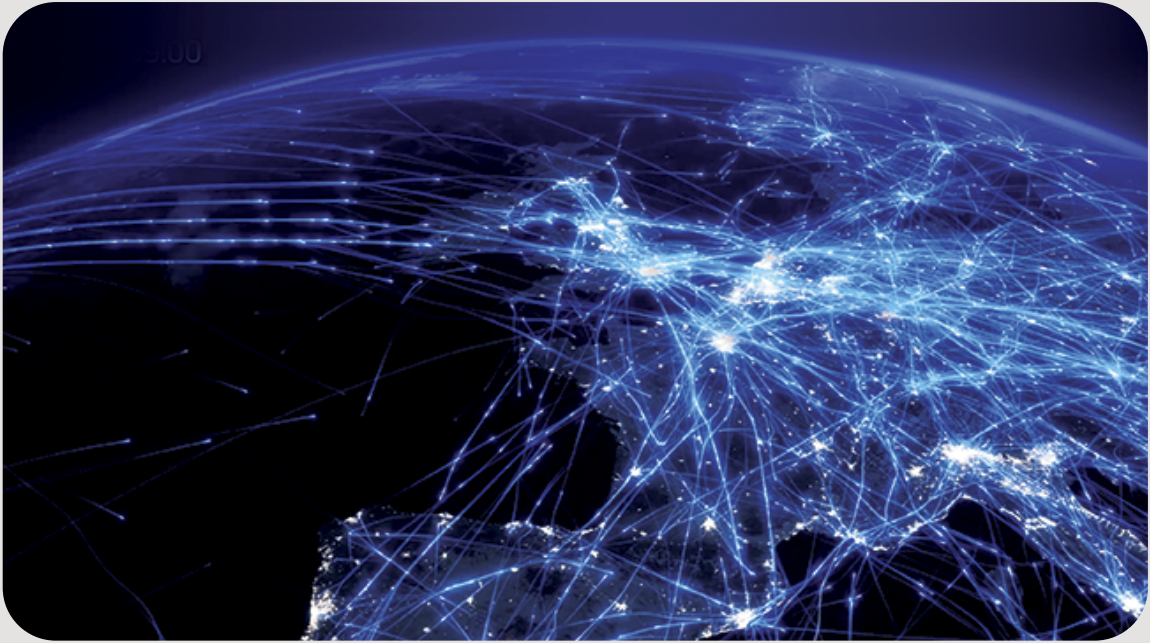

# 1. Introduction

This White Paper sets out to explain the value that metamodelling can bring to air traffic management (ATM) research. It will define metamodelling and explore what it can, and cannot, do. The reader is assumed to have basic knowledge of SESAR: the Single European Sky ATM Research project [1]. An important element of SESAR, as the technological pillar of the Single European Sky initiative, is to bring about improvements, as measured through specific key performance indicators (KPIs), and as implemented by a series of so-called SESAR 'Solutions'. These 'Solutions' are new or improved operational procedures or technologies, designed to meet operational and performance improvements described in the European ATM Master Plan [2].

Central to performance assessment in SESAR is its Performance Framework, and this is supported, in part, by EATMA – the European Air Traffic Management Architecture. This is the common architecture framework for SESAR, its means of integrating operational and technical content developments. In these various SESAR contexts, the term 'metamodel' is not used extensively, and typically describes, at a high level, logical entity relationships, e.g. for performance data and as an architecture mapping and database model. Whilst different definitions of metamodelling indeed prevail in different scientific contexts, usually referring to some form of abstractions of complexity, this White Paper sets out to present a precise definition as deployed in NOSTROMO, relating to simulation metamodels, and the highly specific objectives and corresponding benefits, and limitations, of such application in the performance assessment of SESAR.

Different SESAR Solutions variously deploy different simulations to demonstrate their expected performance contributions across the International Civil Aviation Organisation (ICAO) set of eleven key performance areas (KPAs), using a number of specific KPIs defined in the Performance Framework. Indeed, the corresponding projects are compelled to assess performance expectations as part of the SESAR programme. This brings challenges in terms of computational effort, simulation consistency, assessing KPI interdependencies and general integration.

NOSTROMO does **not set out to build or specify a single, integrated metamodel** for different SESAR Solutions or simulators. Nor does it aim to generalise all such simulations. It is explained in this paper that each simulation metamodel is a modelling proxy for, or simplified abstraction of, a specific Solution (or combination of Solutions) simulation model. Whilst not replacing these simulations, simulation metamodelling is a powerful complementary tool, improving the state of the art for performance assessment, for example in terms of delivering computational efficiency and driving enhanced standardisation.

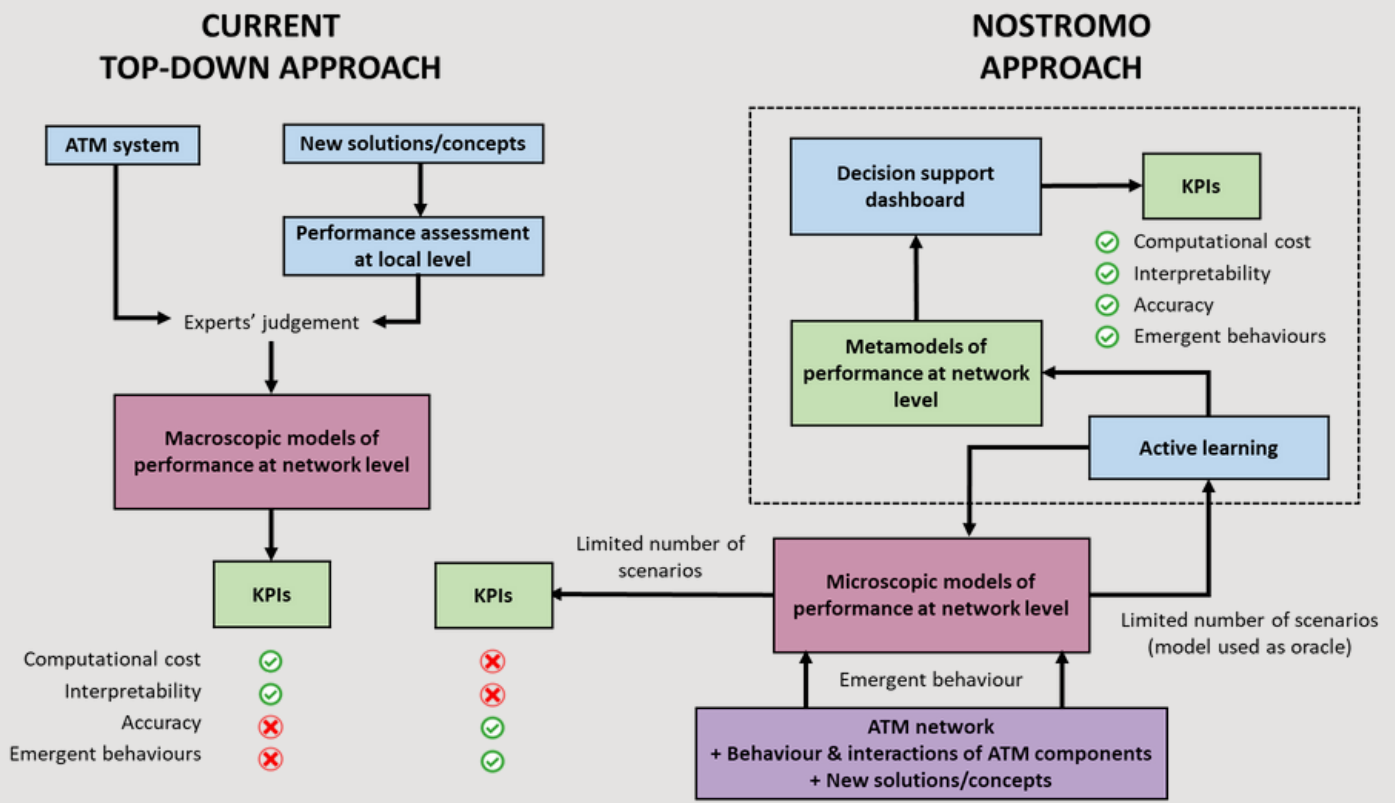

*Figure 1 - Current vs NOSTROMO approach*

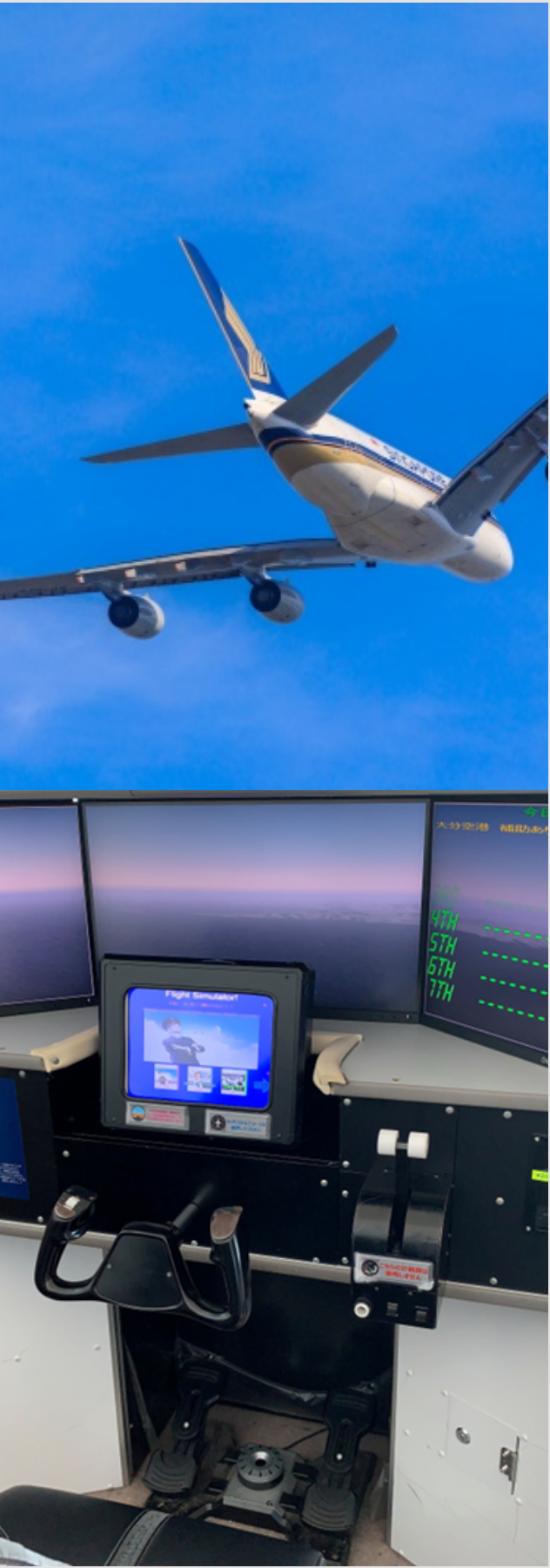

# 2. Definitions

## Simulators

ATM is a complex system where improvements such as SESAR Solutions are constantly proposed to enhance efficiency, capacity, resilience, and sustainability, *inter alia*. Extensive studies are needed to assess the feasibility of the concepts and their potential benefits. Simulators are frequently used to perform these assessments, often across different levels of Solution maturity.

In brief, a simulator is a software program designed to reproduce behaviour likely to occur in the existing (ATM) system. The design of simulators requires efficient and effective computational models for data representation, analysis and visualisation. Various simulator types are available to analysts based on the required level of the investigation and the maturity of the proposed concept:

1. Fast-time simulatros (FTS),
2. Human-in-the-loop simulators (HITL),
3. Real-time simulators (RTS).

## Metamodels

By definition, a **metamodel is a model of a model**. Although the term itself is relatively imprecise, having different meanings and interpretations across the fields where it is used (see [3,4,5], for other SESAR-related metamodels), in this paper, we solely focus on simulation metamodels [6,7,8], that is to say, models specially designed to reproduce the behaviour of simulation models (e.g. simulators).

If a simulation model corresponds to an abstraction of a particular real-world system or phenomenon, a metamodel can be regarded as an abstraction of the simulation model itself, as depicted in figure 2. In this White Paper, we may use the terms 'simulation metamodel' and **'metamodel'** interchangeably; also, we refer to the process of designing and building it as **'metamodelling'**.

Formally, a simulation metamodel is any type of model that can be used to deduce the unknown input-output mapping inherently defined by the simulation model, essentially serving as a **surrogate or proxy** with respect to the associated simulator. Although simulation models are simplified representations of the real-world system, they can still be, and often are, complex and detailed enough to yield significant inconveniences in their use for practical purposes. The most common shortcoming is their tendency to exhibit expensive simulation runs. Furthermore, the size and range of the input variable space can make it difficult to efficiently study and explore the behaviour of computer simulations as a whole, even with current computing technologies.

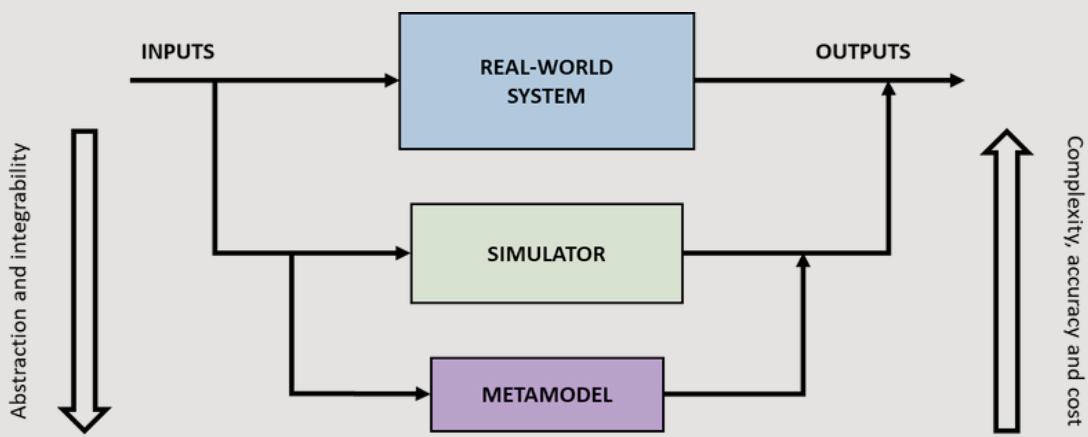

Figure 2 - Relationship between the real-world system under study, the simulator and the metamodel.

Simulation metamodels can then be employed to minimise the computational drawbacks posed by exhausting and time-consuming simulation runs by jointly exploiting their approximate nature, functional simplicity, and fast computing. Being approximations of the underlying simulation functions, the metamodels' design and general performance can achieve balanced trade-offs between computational speed and controlled accuracy loss, depending on their ultimate objectives. Another feature of metamodels is that their respective **functional structures are generally known** and analytically defined, as opposed to those of most simulators. It is worthwhile noting that, although the average arbitrary simulator is often comprised of a plethora of internal analytic expressions and logical relationships, it can be externally treated as a single 'black-box' function with no clear mathematical formula. Nevertheless, an 'emergent behaviour', resulting from its inner interactions and dynamics that evolve over time, can be directly observed. Metamodels aim at mimicking precisely this output behaviour, as a function of the simulation inputs.

Ilustrated in figure 3 is one of the simplest metamodelling scenarios consisting of a simulator with two input and one output variables, along with a simple linear regression model in the role of the metamodel. Here, the metamodeling assumption is that the unknown function **f** represented by the simulator, and consequently its single output, can be reasonably well approximated by a linear combination of its two simulation inputs plus a normally distributed noise term. Naturally, the three parameters of this linear function have to be estimated using some data generated by the simulator itself. This process is typically termed the **'training'** of the model within a machine learning context. In our particular case, this is the process through which the metamodel learns to fit itself to the observed simulation data, ultimately aiming at approximating the simulator's output behaviour.

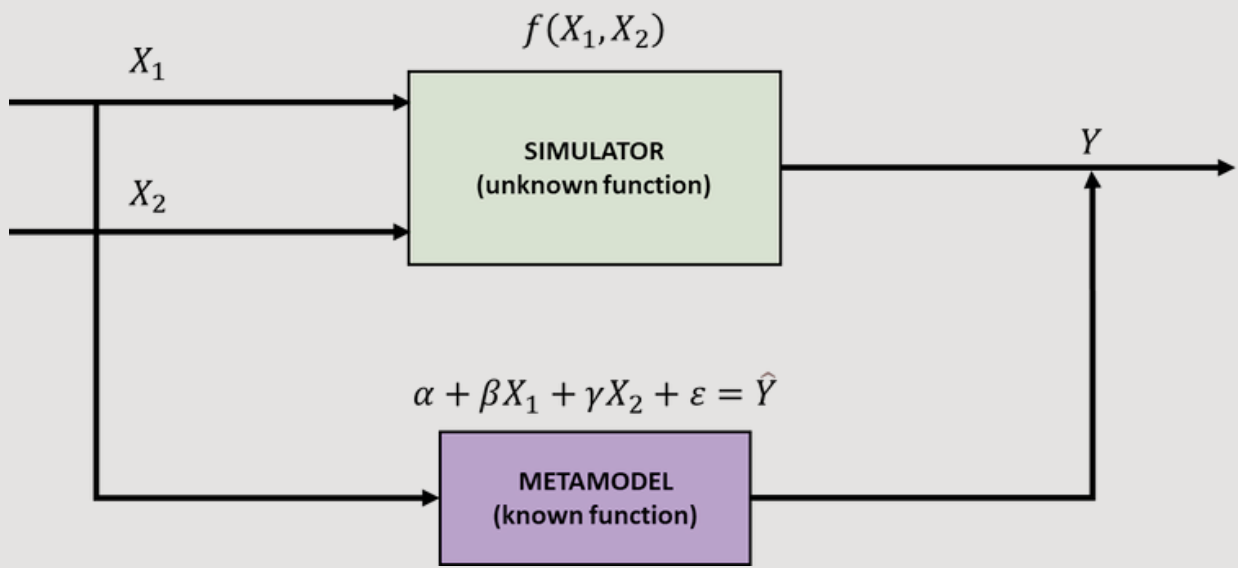

Figure 3 - A simple linear regression model acting as a simulation metamodel.

Despite requiring an initial and unavoidable computational effort, both for sampling the data from the simulator and then for training, the metamodelling approach relies on the fact that most metamodels are (and should be by default) computationally fast, provided that their parameters are already estimated. At this point, if the metamodel represents a fairly good approximation of the simulator, it can thus be employed as a proxy replacement to attain a more efficient exploration of the latter's behaviour. This exploration is conducted by means of predicting the output values for a set (typically a rather large one) of combinations of input values that have not been simulated. Hence, through a surrogate metamodel, exploration by proxy can effectively **bypass the need for new simulation runs** with a minor and controlled accuracy loss and instead generate predictions for unobserved input combinations. Figure 4 summarises two important types of data sets used in metamodeling, namely, the training set to which the metamodel is fitted and the prediction set used to explore the simulation input by proxy and the corresponding output behaviour.

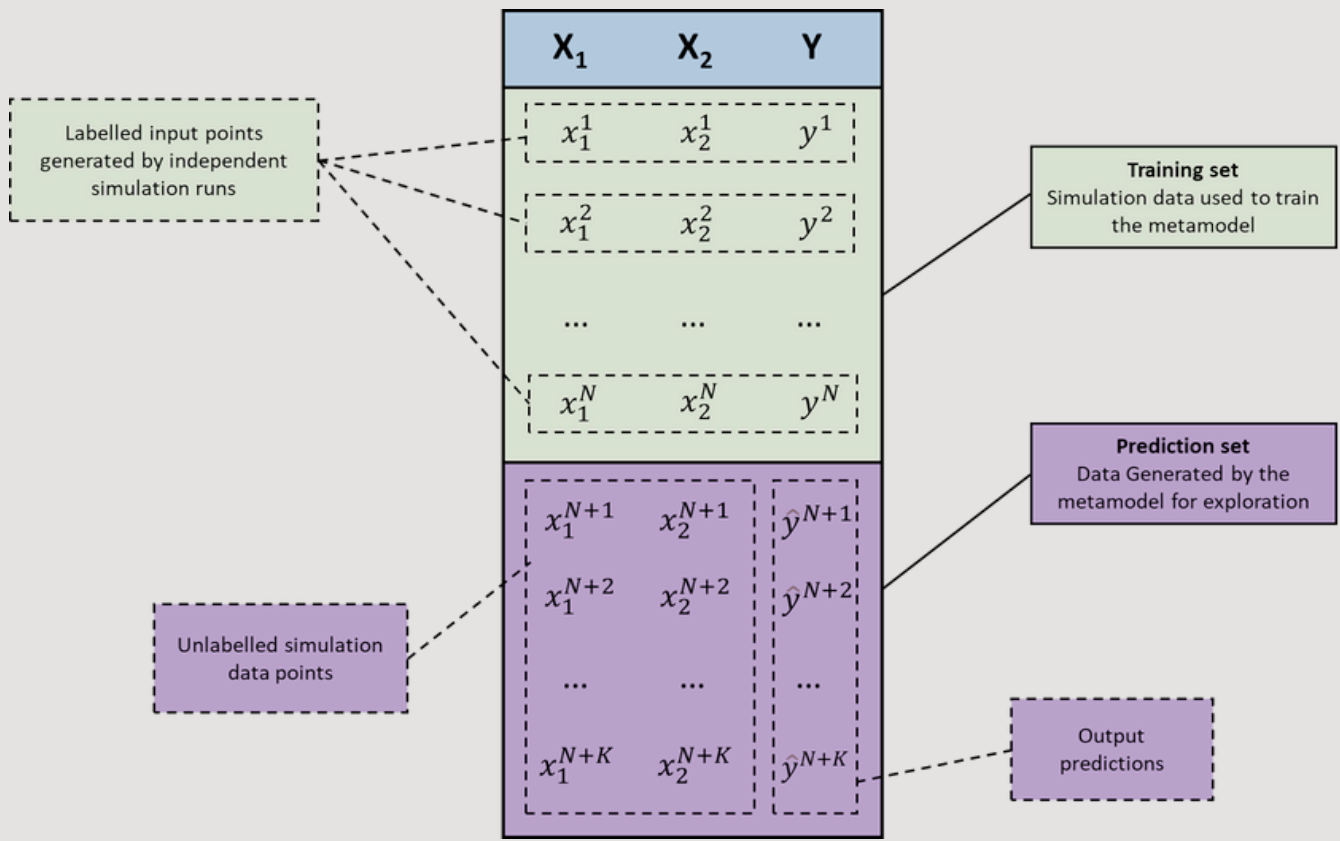

*Figure 4 - Basic ingredients of the training and prediction sets.*

Typically, the training set is called a labelled set, whereas the prediction set in the absence of the predicted values is an unlabelled set. The term 'label', which is commonly associated with classification problems, is here adopted to refer to any value lying in the range of the simulation output space. The exploration process encompasses the prediction of labels which otherwise would have to be generated through simulation, thereby consuming more computational resources and time.

In the context of the NOSTROMO project, a more complex and **powerful family of metamodels is being employed, namely, the Gaussian process (GP)** modelling framework [9,10]. Indeed, GPs have been widely studied and used as simulation metamodels across many different fields, corresponding to the *de facto* default approach in most metamodelling settings. Besides their flexible non-parametric and highly non-linear characteristics, GPs provide a native Bayesian inference system that allows them to handle and quantify their own predictions' uncertainty and the variability naturally present in the data. Compared with other current paradigms, such as Deep Neural Networks, GPs require much less training data. This is a major advantage in simulation settings, because each data point (one simulation) can require substantial resources.

# 3. Active Learning

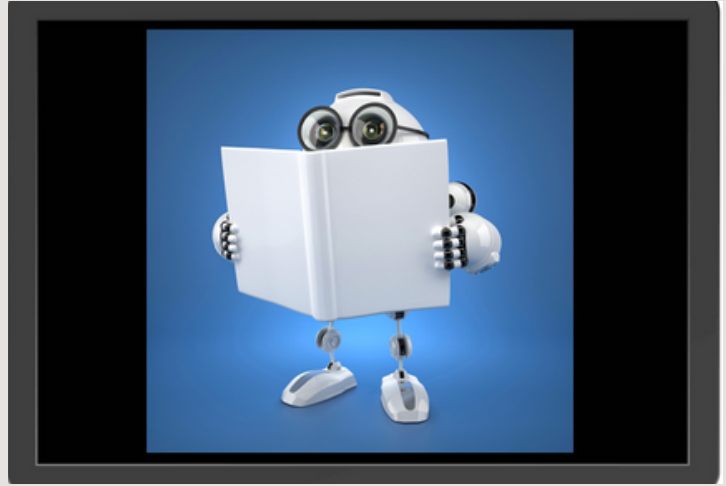

Previously, it was mentioned that metamodels need to be trained with the simulation data so that they can serve as approximators for the simulator at hand. On the other hand, we recognise that simulation results might be computationally expensive and cumbersome to generate in a systematic manner, which ultimately constitutes one of the core issues that metamodelling aims to address.

In this context, **active learning** [11,12] emerges as a powerful learning paradigm that enhances metamodels and underlying algorithms to attain a high predictive performance using as little data as possible. This is achieved by an iterative scheme that, in its simplest form, sequentially selects the most informative input data points to be run through the simulator, eventually adding them to the current training set for model refitting. Here, the informativeness of an arbitrary single unlabelled point is measured as a function of its potential relative contribution to improving the metamodel's performance. In other words, if a data point is associated with a strong information index, then it is more likely to pose a greater performance boost than otherwise. Several information criteria can be adopted, but their reference is out of the scope of this document.

In sum, active learning generally seeks to optimize and, essentially, to accelerate the metamodel's learning curve by avoiding redundancy in the training set, simultaneously making training more efficient and saving significant computational resources, simulation run time, and workload in the process.

Metamodels and active learning are conceptually intertwined and somewhat complementary in practical terms since both generally aim at reducing computational costs. Whereas the metamodels' contribution to this goal lies in providing a parsimonious approximator serving as simulation replacement, **active learning focuses on delivering an efficient training process**. Overall, with the natural combination of the two approaches, more insights concerning the simulator's behaviour are obtained with fewer data, i.e. at a reduced computational cost to the maximum possible extent.

# 4. NOSTROMO integration approach

Within NOSTROMO, the best of both worlds are integrated into a **single auxiliary framework** with the objective of complementing and improving, through **active learning and metamodelling**, the current state of the art for assessing the simulation-supported design and performance impacts of SESAR Solutions on ATM systems. Figure 5 depicts an overview of this architecture's main elements along with its process flow.

Overall, the ultimate goal of this approach is to assist ATM researchers, modellers and practitioners with an auxiliary tool to study the input-output behaviour of simulation models in a more insightful, systematic and computationally efficient fashion. The underlying metamodelling process includes the fulfilment of several prerequisites. First of all, and rather obviously, it cannot advance without a clear selection of the SESAR Solution(s) to be assessed. In addition, these **Solutions must be jointly integrated or implemented** *a priori* into the ATM simulator, upon which metamodelling is then performed.

The simulation model should be capable of encoding the SESAR Solutions into a specific set of input variables and KPIs (output variables) designed for performance assessments. These are the same variables that are eventually used by the metamodel to approximate the simulator's inherent function.

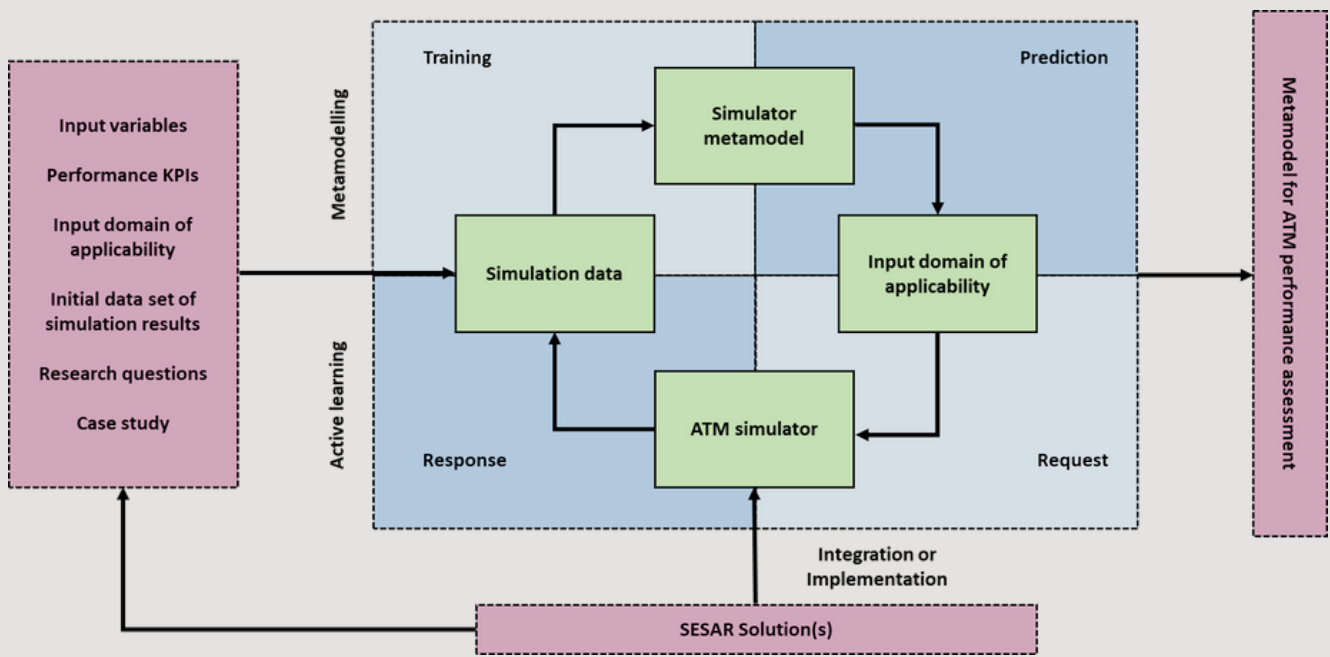

*Figure 5 - Overview of the NOSTROMO's Active learning-based metamodelling architecture.*

Besides the simulation variables of interest, the metamodelling process itself requires the definition of the simulation input region where it should be employed. Here, this region is deemed the input domain of applicability, essentially encompassing the value ranges within which metamodelling is conducted. Furthermore, due to the iterative nature of the approach, an initial (training) set, comprised of previously generated simulation results, must be fed to the metamodel so that the process can be initiated. Finally, it is of utmost importance to keep in mind the research questions to be answered by the metamodel and the case study to be analysed in terms of performance impacts.

Finally, alternating between the metamodeling and the active learning phases, the integrated approach is composed of four elementary steps:

1. **Training**: the metamodel is fitted to the simulation data;

2. **Prediction**: the fitted metamodel is used to predict over the simulation input domain of applicability;

3. **Request**: based on some acquisition criteria, new input data points (unlabelled) are selected to be run by the simulator;

4. **Response**: the simulator provides new simulation output results corresponding to the points from step 3, which are then added to the current training set.

Steps 1-4 are repeated cyclically until a stopping criterion is satisfied. This criterion can be defined, for example, as a function of the metamodel's performance, such as accuracy and error-based metrics, or simply the number of iterations to be performed with respect to the available time, budget and resources. The active learning and metamodelling process eventually provides a trained metamodel designed to help answer the posed research questions and assess the performance impact of the previously selected SESAR Solutions.

It is important to **always bear in mind the approximative nature** of the metamodel which calls for careful handling of the trade-off between speed, accuracy and computational budget. This balance should constantly be monitored and adjusted whenever required, to ensure the metamodeling's ultimate objectives are attained. This means that, if the finally obtained metamodel is not fit for purpose, it can be reintroduced in the active learning metamodelling cycle to allow its parameters to be reestimated. Consequently, and on a similar note, it is equally crucial to recognize and identify the performance threshold from which the mere addition of new training points will not significantly improve the ability of the metamodel to approximate the simulation results. In those cases, and especially from a metamodelling perspective, requesting more simulation results might prove to be a waste of computational resources.

# 5. Applicability of metamodelling approach

## Pros of using the metamodelling approach

As stated previously, a metamodel is a a tool used to approximate the output of a simulator, typically in situations where the simulator is computationally expensive or complex. Characteristics include **functional simplicity, computational speed, and general intelligibility**. Contrary to simulation models, metamodels are explicitly defined by known analytical mathematical formulas, which contributes to an enhanced understanding of the dynamics and associations between the simulation inputs and the outputs of interest. Furthermore, the exploration of the simulation input space, and corresponding output behaviour, is greatly improved. Whilst conducted by surrogate approximation, this exploration allows for fast and efficient identification of patterns and general trends, and it is even able to correct itself via active learning whenever the metamodel's performance starts to drop below unacceptable levels.

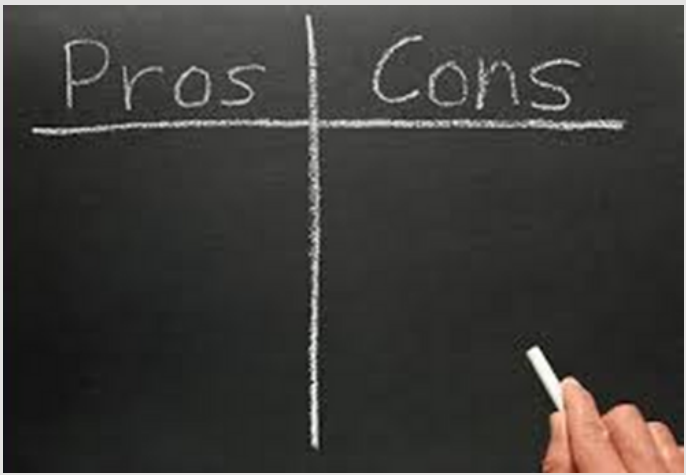

In essence, metamodels serve as stand-ins for simulators. In the context of the SESAR Performance Framework, they are specially conceived to focus on the input/output variables that specifically encode the Solutions under study. Due to their computational speed and their ability to predict 'in bulk', it makes it easy to run up to thousands of combinations of input values in a manner of seconds.

Another important feature of metamodels is their portability. In principle, a fully-trained metamodel should be easily executable and relatively minimal dependencies across different machines and operating systems, especially when compared to the average ATM simulator. For similar reasons, and by adopting current cloud deployment technologies, it should be fairly straightforward to make metamodels available worldwide via application programming interfaces (APIs).

## Cons of using the metamodelling approach

It is essential to be aware of the unavoidable approximate nature of simulation metamodels, which come with a trade-off of decreased accuracy and precision.

Consequently, metamodels should be regarded as auxiliary tools that complement the conventional simulation-based analyses, not their substitution. To this end, NOSTROMO's approach takes advantage of the benefits of simulation metamodelling while also addressing its shortcomings. This approach involves using metamodels and simulation models together, rather than replacing the simulation models, in a combined modelling. While the metamodel aims at reducing the exploratory redundancy by trying to seek the most informative and distinct input data points, the simulation model ensures, by providing labelled data whenever necessary, that this exploration process is maintained close enough to the simulation data distribution.

In practice, another important aspect of metamodelling, especially when coupled with active learning strategies, is that it does **not represent a universal and plug-and-play approach**. Depending on the characteristics and design details of the simulator in question, the construction of corresponding metamodels might require more or less implementation effort. This is particularly true for those cases when the input/output data require some sort of transformation or encoding, for example, from categorical to numerical values or when the simulator runs over multiple data log files. Eventually, **each metamodel is highly tailored for the specific ATM simulator,** SESAR Solutions and case studies under study.

Furthermore, metamodels, like most data-driven models, do also suffer from the so-called 'curse of dimensionality' phenomenon [13,14,15], which refers to the problem of exponential data sparsity in high-dimensional spaces. Whilst metamodeling is a useful approach, it is unwise to consider that a metamodel can entirely approximate the simulation model as a whole. This would be rather impractical, if not virtually impossible, for most simulation approaches with numerous input variables, as the number of required simulation runs would be too high, inevitably rendering the metamodeling itself unattractive and computationally unable to meet its goals. Instead, the **domain of applicability**, or experimental region, should be established first, in which the metamodel should be a valid approximation [7]. In essence, this simplification can be regarded as restricting the metamodel training to a limited area within the sparse simulation input-output space, instead of considering all the input variables at once, which reduces its dimensionality to more manageable and intelligible sizes.

# 6. Metamodel objectives and limitations

## Metamodel capabilities

A metamodel is a tool that quickly and efficiently approximates the output of a simulator. This enables the exploration of the input-output simulation space with much less computational effort. Given a Solution, or multiple ones, already integrated and implemented within the simulation model in question, and a concrete case study, the metamodel is able to run multiple input combination values and predict their corresponding output values in a relatively short amount of time (especially when compared with the simulator's average runtimes), consequently bypassing exhausting and systematic simulations runs. Due to approximation, accuracy is sacrificed to the detriment of faster speeds and exploratory abilities. Besides model 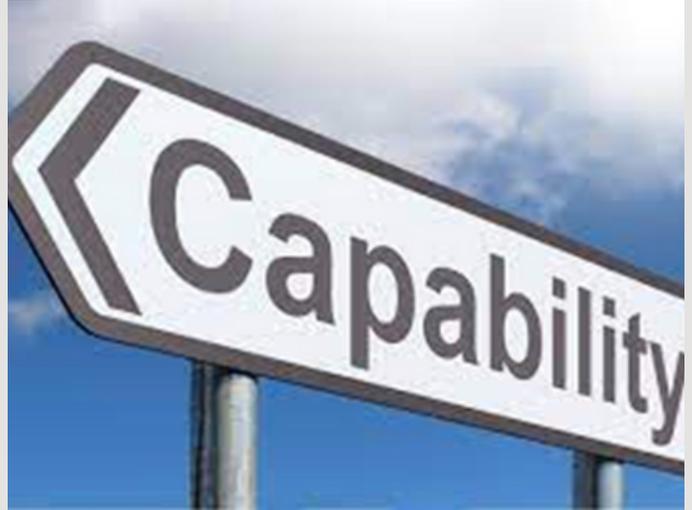 approximation and exploration (prediction) of the simulation input-output mapping, metamodels can also be used for optimization support, sensitivity analysis, and verification and validation [7,13,16,17].

## Limitations of the metamodelling approach

In essence, metamodels act as proxy replacements for simulators. In the context of the SESAR Performance Framework, they are specially conceived to focus on the 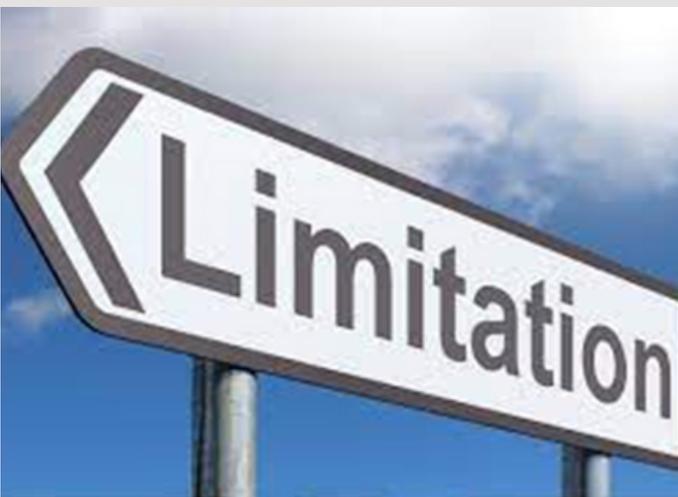 input/output variables that specifically encode the Solutions under study. Therefore, metamodels cannot combine Solutions for themselves, nor they can generalize across different simulation models. Instead, generalization is only conducted across the input space of the same simulator and for a given SESAR Solution. Similarly, metamodels cannot be used for extrapolation purposes for other case studies and sets of input/output variables that have not considered during their design and training in the first place.

# Generalisation of the NOSTROMO approach to any simulator

The metamodel treats the simulation model as a black box that transforms an input space of values into an output space of values, regardless of the numerical computations that run underneath it. Some simulation models are naturally more straightforward to metamodel than others, mostly depending on, but not limited to, the complexity of their input-output relationship, the number and type of variables, and the size or range of the variables' values spaces. Note, however, that whereas the approach is theoretically applicable to any simulator, the obtained metamodels are simulator-specific, thus not generalizable.

# Integration of different Solutions within the metamodelling approach

Most metamodels are agnostic to the design and implementation details of the simulation model it aims to approximate. If the simulator in question already integrates the different Solutions by encoding them into specific simulation rules, function, procedures and input/output variables, then the metamodelling procedure follows naturally, as depicted in figure 6. Therefore, distinct Solutions can be integrated with the metamodeling approach as long as they are first integrated and implemented in the simulator of interest.

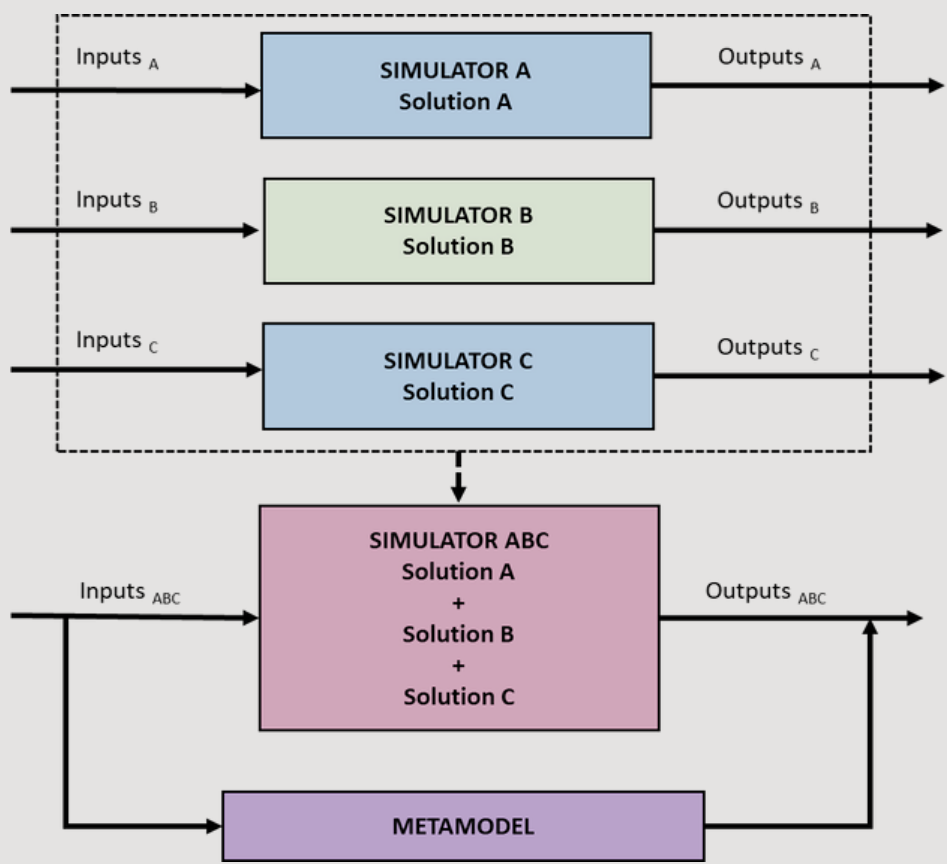

*Figure 6 - Integration of several Solutions into a single simulator for posterior metamodelling.*

The metamodel aims to mimic the latter and not the former. Metamodels are not designed to model Solutions *per se*. Instead, they model the simulation input-output relationships representing Solutions which are encoded into the simulator and represented via simulation input variables and output KPIs.

In the simplest case, a unique metamodel is required per simulator and Solution. However, if Solutions are implemented into a single simulator, the metamodel should be able to accommodate them by adding new input-output variables to the metamodelling dimension space.

## Integration of multiple simulators in a single metamodel

In principle, each metamodel is designed to approximate a specific simulator using a particular set of simulation input-output variables. Metamodels can be regarded as proxy replacements of the underlying simulators, and, as such, they act as computationally faster modelling surrogates. Therefore, metamodels are limited to the simulators for which they were individually built. This means that each metamodel approximately mirrors the behaviour of a given simulation model, especially through generalisation across the simulation input space. Still, it cannot go beyond that, as both are firmly intertwined. Metamodelling represents a single one-to-one relationship between a metamodel and a simulator. Hence, it is **not inherently designed to integrate different simulators**.

In practice, it might be feasible to design a metamodel that integrates multiple simulators insofar that the simulators in question are technically and meaningfully integrable in the first place. Notice, however, that the integration capabilities lie mostly on the integrability of the underlying simulators and not on the metamodelling approach per se. In this case, the metamodel will regard the final integrated simulators simply as a novel simulator.

Figure 7 illustrates two possible simplified ways of integrating two simulators, A and B, with posterior metamodelling in mind, each one individually implementing its own homonymous Solution. The first situation comprises the case where the output of one simulator serves directly as the input to the other.

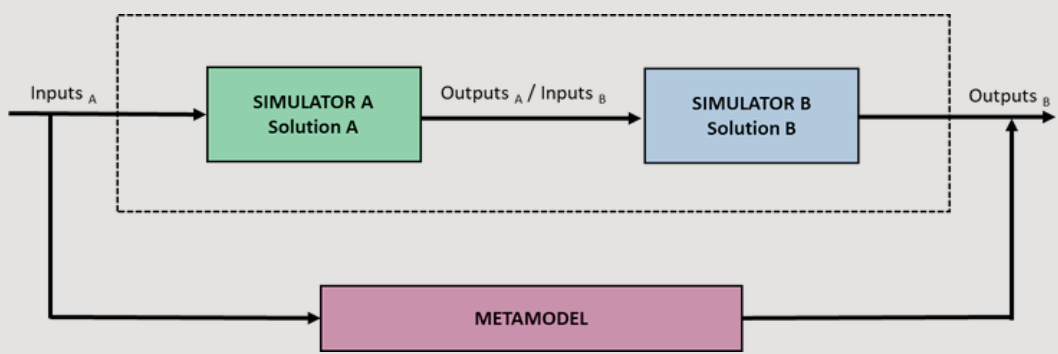

*Figure 7 - Serial integration of two simulators implementing different Solutions and corresponding metamodel.*

Here, we observe that the metamodel approximates the black box whose simulation variables are the inputs of Simulator A and the outputs from Simulator B, being completely agnostic to this serialised integration.

Alternatively, the two simulators can be integrated via parallelisation, as depicted in Figure below. In this situation, both simulators can have their own input spaces and share a set of common variables. The integration itself is performed by somehow individually combining the outputs generated by each simulator in a meaningful and useful way. This integration is independent of and occurs *a priori* to the metamodel's building process. Once again, the metamodel only considers the newly formed simulator's input and output variables, ignoring its inner simulation subcomponents.

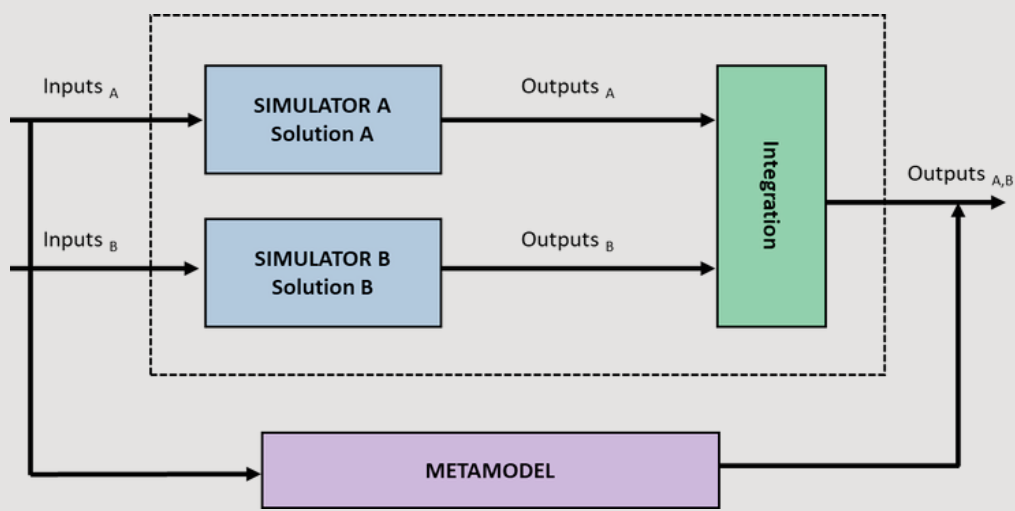

Figure 8 - Parallel integration of two simulators implementing different Solutions and corresponding metamodel.

For all intents and purposes, the simulator resulting from the integration of independent simulators is a new simulator (depicted with dashed lines in both figures) from a metamodelling perspective. The metamodel takes no part in this integration process which is only limited by the ability and utility of combining the simulators in question and, consequently, the associated Solutions. If an arbitrary set of simulators can be integrated in a reasonable and meaningful manner, then a metamodel can be used to approximate the resulting integrated simulator as it is regarded as a new simulation model. Therefore, building a metamodel that encompasses multiple simulators makes sense only if the simulators themselves can be integrated *a priori* and run as a whole. The combination of Solutions, and thus its corresponding metamodelling, is heavily dependent on the success of incorporating them into a single simulator, via integration, as seen before, or from scratch. The lack of standardisation and use of different technologies and programming platforms between simulators might represent a major obstacle in practice. Moreover, even if it is technically possible, one should always firstly investigate if the combination of certain Solutions does indeed make sense from theoretical, research and practical perspectives.

# 7. NOSTROMO dashboard

Metamodels aim to remove computational barriers to perform a complete and efficient exploration of the input-output space defined by complex simulation models. The usefulness of this exploration is ultimately linked to a decision-making process where computational tractability is a necessary but not sufficient condition. The way in which the results of the model are presented is crucial, so that they can be clearly analyzed in order to make better informed decisions.

These two needs can be reconciled through the development of a dashboard equipped with a set of interactive visualisation tools that allow the user to analyse the outputs of the metamodels and explore trade-offs between KPIs with the ultimate purpose of supporting different types of decision-making process related to **performance** management.

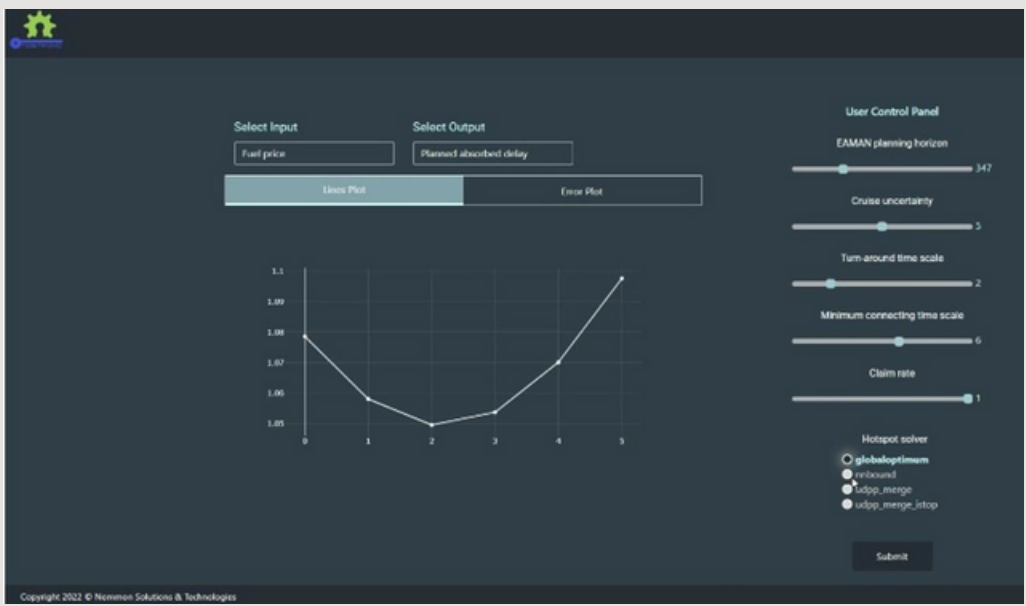

*Figure 8 - NOSTROMO Interactive Dashboard.*

The **NOSTROMO dashboard is a web-based platform** where the user can visualise the impact on performance of alternative decisions in a simple but rigorous way, allowing a comparative assessment. It communicates dynamically with the metamodel to be investigated through the NOSTROMO API to allow a simple and fast exploration of the simulator's input-output space. The methodology chosen for the representation of results is the following: (i) the user can define combinations of inputs of interest through a series of tool selectors; (ii) this information is communicated to the metamodel so that it makes the corresponding prediction; (iii) the metamodel outputs are sent back to the dashboard to be visualised in the chosen plots. Given the computational speed of metamodels, this combined approach enables visual exploration in real time.

# 8. Conclusions and lessons learned

In summary, it is important to note that **metamodels are limited by their intrinsic approximative nature and one-to-one relationship with respect to the simulator being approximated**. NOSTROMO's metamodelling framework is simulator-oriented, i.e. an individual metamodel is produced per simulator, where simulator can be set-up to model one or more Solutions. As already mentioned, metamodels cannot combine Solutions for themselves, nor they can generalize across different simulation models. Instead, generalization is only conducted across the input space of the same simulator and for a given SESAR Solution or Solutions modelled within it. Similarly, metamodels cannot be used for extrapolation purposes for other case studies and sets of input/output variables that have not been considered during their design and training in the first place.

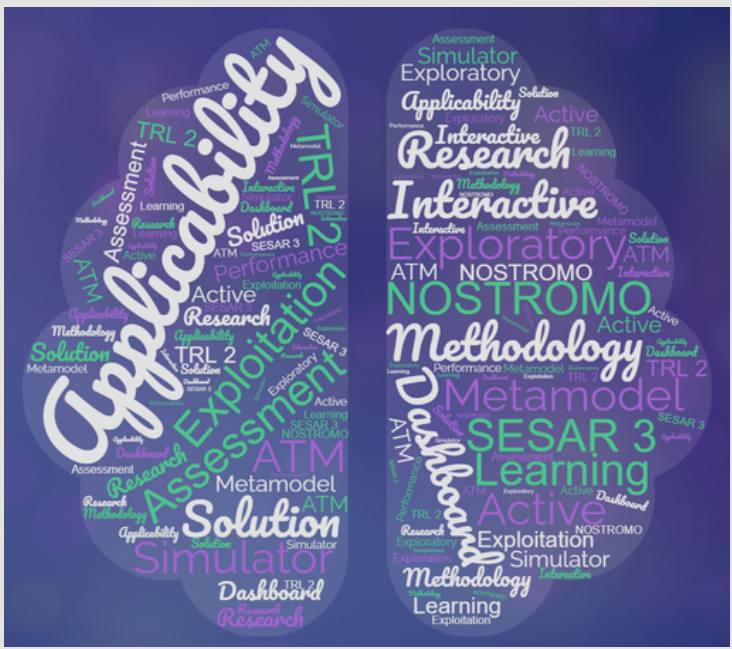

The **effectiveness of metamodelling approach** is largely dependent on the complexity of the simulation model's input-output, the number and type of variables, and the size or range of the variable values spaces. However, the metamodels created are not necessarily generalizable and are specific to the simulation model.

The final results of the project with real case studies and the most complex Solutions selected during the project showed that the metamodeling approach followed by NOSTROMO provides **results very close to the simulator** with much less computational time, allowing a deeper assessment of a Solution, amplifying the exploration of the simulation input and output behaviour space, helping to identify patterns and trends.

The **interactive dashboard built for NOSTROMO** was developed **as a web application**, so that it is accessible from multiple devices by different users concurrently. The system architecture is distributed between the user interface itself (front end) running on the user device and the components executing the system logic (back end) running on a server.

Finally, **additional lessons learned** regard the applicability of the NOSTROMO metamodelling approach to ATM performance assessment:

- There is no universal/plug-and-play or unique metamodeling solution. Each simulation model, case study, and their modelling objectives and research questions often require individual processing and methodological tuning.

- The process of active learning and metamodeling is relatively exploratory. Several parameters, such as the initial data set, stopping criteria, acquisition function and family of metamodels (Gaussian processes, neural networks, etc.) have to be tested before setting them.

- Simulation models with non-numerical input and output variables/parameters require additional steps prior to the application of the methodology itself, such as data conversion and encoding, as well as collection and merging/fusion. The latter is particularly relevant when the simulation data are scattered across multiple log files.

- The wide range of designs and implementations of the simulation models available in the context of SESAR may hinder their compatibility with the current NOSTROMO architecture (methodology + API). While it is true that the developed API constitutes only a proof-of-concept seed of what could become a common SESAR metamodeling platform and novel paradigm for the field, current and future simulation models should be enhanced with their own individual APIs. This should greatly improve the adoption of and integration with the broader and future vision for the proposed architecture.

- The metamodelling coupled with the dashboard effectively make simulation and its results more explainable especially to decision-makers and other stakeholders who are not simulation experts.

- The presented methodology can effectively enhance scenario-based and what-if analyses, greatly contributing to a more comprehensive and in-depth ATM performance assessment framework.

- The NOSTROMO methodology is not meant to replace traditional simulation-based analyses, but to complement them, especially in the area of network-level performance assessment. With that in mind, where the analysis goals lie in the precise assessment of particular micro-level events (e.g. evolution of flight trajectories), metamodelling would not be an appropriate tool.

For various deliverables related to the project, and deeper insights into the specific applications and associated results of NOSTROMO, please refer to the project website (https://nostromo-h2020.eu/).